\documentclass{article}

\usepackage[final]{neurips_2025}

\usepackage[utf8]{inputenc} 
\usepackage[T1]{fontenc}    
\usepackage{hyperref}       
\usepackage{url}            
\usepackage{booktabs}       
\usepackage{amsfonts}       
\usepackage{nicefrac}       
\usepackage{microtype}      
\usepackage{xcolor}         
\usepackage{algorithm}
\usepackage{algpseudocode}
\usepackage{graphicx}
\usepackage{caption}
\usepackage{amsmath}
\usepackage{amsthm}
\usepackage{subcaption} 
\usepackage{array}

\theoremstyle{definition}
\newtheorem{definition}{Definition}

\title{DP-Adam-AC: Privacy-preserving Fine-Tuning of Localizable Language Models Using Adam Optimization with Adaptive Clipping
}

\author{
  Ruoxing (David) Yang \\
  Department of Computer Science \\
  Georgetown University \\
  \texttt{ry216@georgetown.edu} \\
}

\begin{document}

\maketitle

\begin{abstract}

Large language models (LLMs) such as ChatGPT have evolved into powerful and ubiquitous tools. Fine-tuning on small datasets allows LLMs to acquire specialized skills for specific tasks efficiently. Although LLMs provide great utility in both general and task-specific use cases, they are limited by two security-related concerns. First, traditional LLM hardware requirements make them infeasible to run locally on consumer-grade devices. A remote network connection with the LLM provider's server is usually required, making the system vulnerable to network attacks. Second, fine-tuning an LLM for a sensitive task may involve sensitive data. Non-private fine-tuning algorithms produce models vulnerable to training data reproduction attacks. Our work addresses these security concerns by enhancing differentially private optimization algorithms and applying them to fine-tune localizable language models. We introduce adaptable gradient clipping along with other engineering enhancements to the standard DP-Adam optimizer to create DP-Adam-AC. We use our optimizer to fine-tune examples of two localizable LLM designs, small language model (Qwen2.5-0.5B) and 1.58 bit quantization (Bitnet-b1.58-2B). We demonstrate promising improvements in loss through experimentation with two synthetic datasets. 

\end{abstract}

\section{Introduction}
Large language models (LLMs) such as ChatGPT have evolved into powerful tools used ubiquitously in professional and casual settings. However, fully harnessing the capabilities of an LLM and optimizing it for a specific task often requires some fine-tuning on domain-specific data. For example, an LLM-based email assistant optimized for a particular user will benefit greatly from fine-tuning on that particular user's past email exchange data. 

\paragraph{Security concerns.} Given its powerful utility, users may want to deploy fine-tuned task-specific LLMs in hostile, high-surveillance environments such as an authoritarian state. An example user operating in a hostile foreign regime may benefit from an LLM assistant fine-tuned on sensitive military data that can provide fast, informed military advice. Traditional LLM fine-tuning frameworks render this deployment scenario infeasible due to two main security concerns. 

\paragraph{1. Training Data Leakage.} Both traditional and fine-tuned LLMs are vulnerable to data reconstruction attacks, where an attacker can reproduce training data from a trained model \cite{du2025privacyfinetuninglargelanguage, liu2025datareconstructionattacksdefenses}. In our scenario, if the hostile regime gains access to the model, they may be able to reproduce and access sensitive data. 

\paragraph{2. Hardware Limitations.} Traditional LLMs require powerful and dedicated hardware during both training and inference. An LLM end-user can usually only access the tool over a remote network connection to a dedicated LLM server. An attacker may eavesdrop on the connection and gain access to sensitive LLM prompts and output. While traditional network security measures such as end-to-end encryption still apply in this scenario, the most secure solution is to discard the network connection and host the language model locally. 

\paragraph{Setting.} We address these security concerns by applying differentially private fine-tuning to two localizable LLM designs. Small language models (SLM) provide one solution to the localization problem by greatly reducing the amount of parameters used by the model while preserving utility \cite{belcak2025smalllanguagemodelsfuture}. Models with order of magnitude fewer parameters require much less computational resources to operate and can feasibly be adapted to local devices. We also study the BitNet-b1.58 language model, which uses extreme ternary-bit quantization to reduce the computational intensity of LLM inference while preserving a high parameter count and performance. 

To fine-tune these models, we adapt and enhance current differentially private optimization methods to introduce DP-Adam-AC. This optimizer is based on DP-AdamW \cite{sun2025dpadamw} and incorporates a variation of adaptive clipping \cite{du2022dynamicdifferentialprivacypreservingsgd} as well as several other engineering enhancements, such as exponential-moving-average (EMA) smoothing at training evaluation time and dynamic clip-rate based learning-rate adjustment. We demonstrate promising performance on fine-tuning experiments on two synthetic datasets. 

\paragraph{Contribution.} Our work contributes both theoretical empirical insights to differentially private fine-tuning of localizable language models. First, we introduce DP-Adam-AC, a modified version of the DP-AdamW optimizer specialized for fine-tuning language models. We also provide an implementation of the Renyi Differential Privacy (RDP) \cite{mironov2019renyidifferentialprivacysampled} privacy accountant with variable q. Last, we present an empirical analysis of training loss and inference resource consumption of our proposed framework to investigate the utility and feasibility of LLM deployment in security-sensitive scenarios. 

\paragraph{Availability.} All software implementations from this project are available as open-source resources, see Appendix~\ref{appendix-code}.

\section{Related Work}

\subsection{Differentially Private Optimization}

The classic approach to privacy in optimization is differentially private stochastic gradient descent (DP-SGD), which performs standard gradient descent with clipped gradients that include parameterized gaussian noise\cite{Abadi_2016}. Under such scenarios, privacy loss is usually tracked with Rényi Differential Privacy (RDP) accountants \cite{mironov2019renyidifferentialprivacysampled}. While DP-SGD is empirically effective on training small models to perform standard machine learning tasks such as classification, it's simple nature and fixed clipping design restricts its utility on larger models such as LLMs.  

To mitigate over/under-clipping issues introduced by the fixed clip-rate of classic DP-SGD, \cite{du2022dynamicdifferentialprivacypreservingsgd} introduced a class of Dynamic DP-SGD optimizers that adjust the clipping rate during training, improving training convergence stability. This work also presents a differentially private Adam optimizer with adaptive clipping. Our DP-Adam-AC adopts the principle of adaptive clipping but modifies it by adjusting the clip rate based on historic clipping activation statistics instead of a fixed decay curve. We also add other engineering mechanisms such as EMA-based evaluation smoothing and clip-rate-driven learning-rate adjustment. 

Other studies have also investigated differentially private Adam-style optimizers. \cite{tang2023dpadambcdpadamactuallydpsgd} showed that DP-Adam without bias correction can reduce to DP-SGD, highlighting the importance of bias correction. \cite{sun2025dpadamw} investigated DP-AdamW, which updates DP-Adam by introducing decoupled weight decay and analyzed its privacy/utility trade-offs. From a different angle, \cite{charles2024finetuninglargelanguagemodels} studied example- and user-level DP in standard LLM fine-tuning, demonstrating feasibility at scale. Complementary regularization-based approaches such as DP-Flat encourage flatter loss landscapes to improve privacy-preserving fine-tuning \cite{chen2024privacypreservingfinetuninglargelanguage}. However, none of these approaches considers resource constraints and model localization. 

\subsection{Attacks on Non-private Language Models}

Non-private LLMs are vulnerable to data reconstruction and extraction attacks that can surface sensitive training records from model parameters or sampled text. \cite{liu2025datareconstructionattacksdefenses} provide a systematic evaluation showing heightened risk with larger models and memorized content. \cite{du2025privacyfinetuninglargelanguage} survey privacy risks specific to fine-tuning workflows and summarize defenses. These findings motivate us to apply formal DP during LLM training and fine-tuning.

\subsection{Compact, Localizable Language Models}

Local deployment reduces reliance on remote server connections and the associated network attack interface, but it demands efficient models that still provide utility. Small language models (SLMs) significantly reduce parameter counts while retaining strong utility, improving feasibility on consumer hardware \cite{belcak2025smalllanguagemodelsfuture}. On the other hand, extreme quantization, such as seen in BitNet-b1.58, achieves fast, low-memory CPU inference with ternary weights \cite{wang20241bitaiinfra11}. Each arithmetic operation with ternary weights can be reduced to either zeroing out the parameter or adjusting its sign (multiplying by 0, -1, or 1). This simpler computational task not only reduces the costs of inference but also leaves great room for future hardware optimizations. We evaluate DP-Adam-AC on both SLM and BitNet style architectures and demonstrate the practicality of privacy-preserving fine-tuning on localizable models.

\section{Preliminaries}
\subsection{Optimization}

Optimization algorithms play a central role in training and fine-tuning large language models. In this section we briefly review stochastic gradient descent (SGD), Adam, and AdamW, the most popular modern optimizers. We provide pseudocode for these optimizers in Appendix~\ref{appendix-optimizers}.

\paragraph{Stochastic Gradient Descent (SGD).}
SGD iteratively updates parameters in the opposite direction of the gradient of the loss function with respect to the parameters:
\begin{equation}
    \theta_{t+1} = \theta_t - \eta_t \, g_t,
\end{equation}
where $g_t = \nabla_{\theta} \ell(\theta_t; x_t)$ is the stochastic gradient evaluated on a mini-batch $x_t$, and $\eta_t$ is the learning rate.  
A common enhancement is \emph{momentum}, which accumulates an exponentially decaying moving average of past gradients:
\begin{align}
    m_t &= \beta m_{t-1} + (1 - \beta) g_t, \\
    \theta_{t+1} &= \theta_t - \eta_t \, m_t,
\end{align}
where $\beta \in [0,1)$ controls the influence of past gradients. Momentum helps damp oscillations and accelerate convergence in valleys of the loss surface.

\paragraph{Adam.}
Adam \cite{kingma2017adammethodstochasticoptimization} extends SGD with per-parameter adaptive learning rates and two moment estimates:  
a first moment (mean of gradients) $m_t$ and a second moment (uncentered variance) $v_t$:
\begin{align}
    m_t &= \beta_1 m_{t-1} + (1 - \beta_1) g_t, \\
    v_t &= \beta_2 v_{t-1} + (1 - \beta_2) g_t^2,
\end{align}
where the squares are elementwise and $\beta_1,\beta_2 \in [0,1)$ control the exponential moving averages.  
To correct initialization bias, Adam uses:
\begin{align}
    \hat{m}_t &= \frac{m_t}{1 - \beta_1^t}, \\
    \hat{v}_t &= \frac{v_t}{1 - \beta_2^t}.
\end{align}
The update is:
\begin{equation}
    \theta_{t+1} = \theta_t - \eta \frac{\hat{m}_t}{\sqrt{\hat{v}_t} + \epsilon},
\end{equation}
with $\epsilon > 0$ ensuring numerical stability.  

\paragraph{AdamW.}
AdamW \cite{loshchilov2019decoupledweightdecayregularization} modifies Adam by decoupling weight decay from the gradient update, addressing over-regularization issues in Adam with $\ell_2$ regularization.  
The update rule becomes:
\begin{equation}
    \theta_t \gets \theta_t - \eta \lambda \theta_t, \quad
    \theta_{t+1} \gets \theta_t - \eta \frac{\hat{m}_t}{\sqrt{\hat{v}_t} + \epsilon},
\end{equation}
where $\lambda$ is the weight decay coefficient applied directly to parameters before the adaptive step.  
This decoupling leads to better generalization in many large-scale training scenarios.

In this work, we build on the Adam framework, adapting it for differentially private fine-tuning by introducing adaptive clipping and additional mechanisms.

\subsection{Differential Privacy}

Differential Privacy (DP) \cite{dwork2006calibrating} provides a formal mathematical framework for quantifying and limiting the information an algorithm can leak about any individual data record in its input. These are the standard definitions for differential privacy.

\begin{definition}[$(\varepsilon,\delta)$-Differential Privacy]
A randomized mechanism $\mathcal{M} : \mathcal{X}^n \rightarrow \mathcal{Y}$ satisfies $(\varepsilon,\delta)$-differential privacy if for all pairs of adjacent datasets $D, D' \in \mathcal{X}^n$ differing in at most one record, and for all measurable subsets $S \subseteq \mathcal{Y}$:
\begin{equation}
    \Pr[\mathcal{M}(D) \in S] \le e^{\varepsilon} \Pr[\mathcal{M}(D') \in S] + \delta.
\end{equation}
\end{definition}
Here, $\varepsilon \ge 0$ is the \emph{privacy loss} parameter and $\delta \in [0,1]$ is the probability of violating pure $\varepsilon$-DP. Smaller $\varepsilon$ and $\delta$ correspond to stronger privacy guarantees.

\paragraph{Sensitivity and the Gaussian Mechanism.}
For a function $f : \mathcal{X}^n \to \mathbb{R}^d$, its $\ell_2$-sensitivity is:
\begin{equation}
    \Delta_2 f = \max_{\substack{D,D' \text{ adjacent}}} \| f(D) - f(D') \|_2.
\end{equation}
The Gaussian Mechanism achieves $(\varepsilon,\delta)$-DP by adding Gaussian noise $\mathcal{N}(0, \sigma^2 I_d)$ with:
\begin{equation}
    \sigma \ge \frac{\Delta_2 f \sqrt{2 \ln (1.25/\delta)}}{\varepsilon}.
\end{equation}

\paragraph{Rényi Differential Privacy (RDP).}
Rényi Differential Privacy \cite{mironov2019renyidifferentialprivacysampled} offers a more flexible composition analysis by measuring the divergence between output distributions at a fixed order $\alpha > 1$.

\begin{definition}[$(\alpha,\varepsilon)$-RDP]
A mechanism $\mathcal{M}$ satisfies $(\alpha,\varepsilon)$-Rényi Differential Privacy if for all adjacent $D, D'$:
\begin{equation}
    D_{\alpha}\!\left( \mathcal{M}(D) \,\|\, \mathcal{M}(D') \right) \le \varepsilon,
\end{equation}
where $D_{\alpha}(P \| Q)$ is the Rényi divergence of order $\alpha$:
\begin{equation}
    D_{\alpha}(P \| Q) = \frac{1}{\alpha - 1} \log \mathbb{E}_{x \sim Q} \left[ \left( \frac{P(x)}{Q(x)} \right)^{\alpha} \right].
\end{equation}
\end{definition}
RDP composes additively across multiple mechanism invocations:
\begin{equation}
    (\alpha, \varepsilon_1)\text{-RDP} \;+\; (\alpha, \varepsilon_2)\text{-RDP} \;\Rightarrow\; (\alpha, \varepsilon_1 + \varepsilon_2)\text{-RDP}.
\end{equation}
An $(\alpha,\varepsilon)$-RDP guarantee can be converted into an $(\varepsilon',\delta)$-DP guarantee for any $\delta > 0$ via:
\begin{equation}
    \varepsilon' = \varepsilon + \frac{\log(1/\delta)}{\alpha - 1}.
\end{equation}

\paragraph{Subsampled Gaussian Mechanism.}
In deep learning, DP-SGD \cite{Abadi_2016} uses gradient clipping to bound sensitivity, followed by adding Gaussian noise to the clipped gradients. The subsampled Gaussian mechanism, where each update uses a random mini-batch, is well-analyzed under RDP \cite{mironov2019renyidifferentialprivacysampled}, enabling tight privacy accounting for many iterations.

\section{Technical Approach}

In this section, we present the technical details and design choices of the DP-Adam-AC algorithm. We also describe our implementation of a variable-q RDP privacy accountant. Finally, we briefly describe the workflow of our proposed framework. 

\subsection{Fine-Tuning Optimization Algorithm}

DP-Adam-AC introduces to the standard family of differentially private Adam-based optimizers three main engineering enhancements: history-based adaptive clipping, exponential-moving-average (EMA) smoothing, and clip-based dynamic learning-rate. These design choices improve training stability under noisy conditions. We present our full algorithm in Section~\ref{section-algorithm}.

\paragraph{Micro-batching and DP-Noise.} Traditional DP-SGD clips the gradient of each training example individually before adding noise to ensure the influence of each gradient is controlled. This allows the formal privacy guarantees of DP-SGD to hold. Processing each gradient individually is unreasonable for our framework due to memory constraints for large models such as LLMs. We use micro-batching to resolve this issue. We split the training data batch into smaller micro-batches and compute the gradient for the micro-batch. We clip these gradients based on our adaptive clipping mechanism and accumulate them into a sum. Finally, we add parameterized DP Gaussian noise to the gradient sum, then average over the total batch size to retrieve the privatized individual gradient for the micro-batch. We provide pseudocode for micro-batching in Appendix~\ref{appendix-micro-batching}.

\paragraph{Adaptive Clipping.} 
\cite{du2022dynamicdifferentialprivacypreservingsgd} adapts the gradient clipping rate at a fixed parameterized schedule. This approach assumes that gradients shrink predictably during the training process as loss drops and accordingly weakens clipping to enhance stability. This works with simpler models and tasks; however, fine-tuning an LLM not only involves larger models with more parameters but also inherently entails a shorter training period. Fine-tuning is performed after the main training phase of an LLM to enhance LLM specialization and involves shorter training schedules. Shorter schedules, larger training data sizes, and large parameter counts weaken the gradient-shrinkage assumption.

We adapt the gradient clipping norm $C$ during training to maintain a target clip rate $\rho^\star$ (e.g., $\rho^\star = 0.20$), defined as the fraction of microbatches whose pre-clip gradient norms exceed their clipping threshold.

For each microbatch $i$ of size $m_i$, let $n_i$ be its pre-clip global $\ell_2$-norm:
\[
n_i \;=\; \big\| \nabla_\theta \ell(X_i) \big\|_2,
\]
where $\ell(X_i)$ is the loss over that microbatch. We define its \emph{unit norm} as
\[
u_i \;=\; \frac{n_i}{\max(1, m_i)}.
\]

We maintain a history buffer $\mathcal{U}$ containing the most recent $H$ unit-norm values. At each step, we compute the target percentile
\[
q \;=\; 100 \cdot \big( 1 - \rho^\star \big),
\]
and set the clipping norm to
\[
C \;\leftarrow\; \mathrm{clip}\!\big(P_q(\mathcal{U}),\; C_{\min},\; C_{\max}\big),
\] 
where $P_q(\mathcal{U})$ is the $q$-th percentile of $\mathcal{U}$ and $\mathrm{clip}(x, a, b) = \min(b, \max(a, x))$.

This procedure keeps the empirical clip rate close to $\rho^\star$ while automatically adjusting to changing gradient scales during training.

\paragraph{Exponential Moving Average (EMA) Smoothing for Evaluation.}

To reduce the variance introduced by differential privacy noise during training, we maintain an exponential moving average (EMA) of the model parameters for evaluation. This EMA model is updated after each optimization step but is not used for gradient computation.

Let $\theta_t$ be the model parameters at step $t$ and $\hat{\theta}_t$ be the EMA parameters. The EMA update is:
\[
\hat{\theta}_t \;=\; d \cdot \hat{\theta}_{t-1} \;+\; (1-d) \cdot \theta_t,
\]
where $d \in (0, 1)$ is the decay factor (e.g., $d = 0.999$).

During evaluation, we temporarily replace $\theta_t$ with $\hat{\theta}_t$:
\[
\theta_{\mathrm{eval}} \;=\; \hat{\theta}_t,
\]
run the forward pass to compute metrics, and then restore $\theta_t$ to continue training.

Since EMA is a deterministic post-processing of the noisy parameters, it does not incur any additional privacy cost. The EMA model is the final outputted model.

\paragraph{Clip-based Dynamic Learning-rate.}

We adjust the learning rate multiplier $\gamma$ during training based on the observed clip rate $\rho$, defined as the fraction of microbatches whose pre-clip gradient norm exceeds the clipping threshold:
\[
\rho \;=\; \frac{\#\{\, i \mid n_i > C \cdot m_i \,\}}{\#\text{microbatches}},
\]
where $n_i$ is the pre-clip global $\ell_2$-norm for microbatch $i$ of size $m_i$, and $C$ is the current clipping norm.

Two thresholds, $\rho_{\mathrm{low}}$ and $\rho_{\mathrm{high}}$, define a target operating range for the clip rate. We update $\gamma$ as:
\[
\gamma \;\leftarrow\;
\begin{cases}
\min\!\left(\gamma_{\max},\ \gamma \cdot \uparrow \right),
    & \rho < \rho_{\mathrm{low}}, \\[4pt]
\max\!\left(\gamma_{\min},\ \gamma \cdot \downarrow \right),
    & \rho > \rho_{\mathrm{high}}, \\[4pt]
\gamma, & \text{otherwise},
\end{cases}
\]
where $\uparrow > 1$ and $\downarrow < 1$ are gentle multiplicative factors (e.g., $\uparrow = 1.01$, $\downarrow = 0.995$), and $[\gamma_{\min},\gamma_{\max}]$ bounds the allowed multiplier range.

The effective learning rate at step $t$ becomes $\eta_t = \gamma \cdot \mathrm{lr\_base}(t)$.

where $\mathrm{lr\_base}(t)$ is the underlying learning-rate schedule (e.g., cosine decay with warmup). This mechanism keeps training within a balanced clipping regime, improving stability and convergence under noisy DP optimization.

\textbf{Other Design Choices.} In classic optimization, AdamW (with decoupled weight decay) consistently outperforms Adam; however, our design intentionally avoids weight decoupling. In differentially private training, the injected Gaussian noise already acts as a strong implicit regularizer by perturbing gradients at every step. Adding decoupled weight decay on top of this can overly constrain parameter magnitudes, leading to underfitting. Discarding weight decay allows the model to use its full representational power while still benefiting from the regularization effect of DP noise. 

\subsection{Variable-q RDP Accountant}

We track privacy loss under the subsampled Gaussian mechanism with a per-step sampling probability $q_t = \tfrac{B_t}{N}$.
For Rényi order $\alpha \in \{2,3,\dots\}$, the step-wise RDP of Poisson/batch
subsampling with Gaussian noise $\mathcal{N}(0,\sigma^2 I)$ admits the tight
integer-order bound \cite{mironov2019renyidifferentialprivacysampled}:
\[
\mathrm{RDP}_\alpha(q,\sigma)
\;=\;
\frac{1}{\alpha-1}\,
\log\!\left(
\sum_{j=0}^{\alpha}
\binom{\alpha}{j}\,
q^{\,j}\,(1-q)^{\alpha-j}\,
\exp\!\Big(\tfrac{j(j-1)}{2\sigma^2}\Big)
\right).
\]

Across training steps $t=1,\dots,T$, RDP composes additively:
\[
\mathrm{RDP}^{\mathrm{tot}}_\alpha
\;=\;
\sum_{t=1}^{T} \mathrm{RDP}_\alpha(q_t,\sigma).
\]
We then convert to $(\varepsilon,\delta)$-DP via
\[
\varepsilon(\delta)
\;=\;
\min_{\alpha \in \mathcal{A}}
\left\{
\mathrm{RDP}^{\mathrm{tot}}_\alpha
\;+\;
\frac{\log(1/\delta)}{\alpha-1}
\right\},
\]
where $\mathcal{A}$ is a finite set of integers.

Using the exact $q_t$ each step (variable-q) tightens the account over assuming a fixed sampling rate. All calculations are performed in log-space to avoid overflow. Pseudocode for the accountant is provided in Appendix~\ref{appendix-accountant}.

\subsection{Workflow Environment}

Within our framework, we expect the fine-tuning process to be completed at a facility with powerful computational resources such as GPUs. After fine-tuning is complete, the mature model can be exported and deployed on localized, consumer-grade (CPU) hardware, and transported to sensitive environments to operate in a self-sustained manner.

\section{DP-Adam-AC Algorithm}
\label{section-algorithm}

\begin{algorithm}[H]
\caption{DP-Adam-AC}
\label{alg:dp_adam_ac}
\begin{algorithmic}[1]
\Require $f_\theta$, batch $X$ of size $B$, microbatch size $m_{\text{micro}}$, $\mathrm{lr\_base}(t)$, $\sigma$, $C$, $\rho^\star$, $H$, $(\beta_1,\beta_2,\epsilon)$, $d$, $\gamma_{\min},\gamma_{\max}$, $\uparrow,\downarrow$, $\rho_{\text{low}},\rho_{\text{high}}$
\State $(m,v)\leftarrow(0,0)$, $\gamma \leftarrow 1$, $\hat\theta\leftarrow\theta$, $\mathcal{U}\leftarrow[~]$
\For{$t=1,2,\dots$}
  \State $\mathrm{SumGrad}\leftarrow 0$, $\mathrm{PreNorms}\leftarrow[~]$, $\mathrm{Sizes}\leftarrow[~]$
  \For{$X_i$ in microbatches of $X$}
    \State $\ell \leftarrow \mathrm{Loss}(f_\theta(X_i), X_i)$
    \State $\nabla \leftarrow \nabla_\theta \ell$, $n_i \leftarrow \|\nabla\|_2$
    \State $\nabla \leftarrow \nabla \cdot \min\!\big(1, \tfrac{C|X_i|}{n_i}\big)$
    \State $\mathrm{SumGrad} \leftarrow \mathrm{SumGrad} + \nabla$
    \State append $n_i$ to $\mathrm{PreNorms}$, $|X_i|$ to $\mathrm{Sizes}$
    \State zero gradients
  \EndFor
  \State $\mathrm{SumGrad} \leftarrow \mathrm{SumGrad} + \mathcal{N}(0,(\sigma C)^2 I)$
  \State $g \leftarrow \mathrm{SumGrad} / B$
  \State $m \leftarrow \beta_1 m + (1-\beta_1) g$, $v \leftarrow \beta_2 v + (1-\beta_2)(g\odot g)$
  \State $\hat m \leftarrow m/(1-\beta_1^t)$, $\hat v \leftarrow v/(1-\beta_2^t)$
  \State $\theta \leftarrow \theta - \gamma\,\mathrm{lr\_base}(t)\,\hat m / (\sqrt{\hat v}+\epsilon)$
  \State $\hat\theta \leftarrow d\,\hat\theta + (1-d)\,\theta$ \Comment{EMA update for evaluation}
  \State $\rho \leftarrow \frac{1}{|\mathrm{Sizes}|}\sum_i \mathbf{1}\{\,\mathrm{PreNorms}_i > C\,\mathrm{Sizes}_i\,\}$
  \State append $\mathrm{PreNorms}_i/\max(1,\mathrm{Sizes}_i)$ to $\mathcal{U}$; keep last $H$
  \State $q \leftarrow 100(1-\rho^\star)$
  \State $C \leftarrow \mathrm{clip}(P_q(\mathcal{U}),\,C_{\min},\,C_{\max})$ \Comment{Adaptive clipping update}
  \If{$\rho < \rho_{\text{low}}$} \State $\gamma \leftarrow \min(\gamma_{\max},\,\gamma\,\uparrow)$ \Comment{LR nudge up}
  \ElsIf{$\rho > \rho_{\text{high}}$} \State $\gamma \leftarrow \max(\gamma_{\min},\,\gamma\,\downarrow)$ \Comment{LR nudge down}
  \EndIf
\EndFor
\end{algorithmic}
\end{algorithm}

\section{Experiment Setup}

\paragraph{Overview:} To verify the performance of DP-Adam-AC on fine-tuning localizable language models, we compare the loss reduction performance of fine-tuning using DP-Adam-AC across different privacy settings (including non-private fine-tuning) with baseline non-private versions of an SLM model and a BitNet model. We choose Qwen2.5-0.5B \cite{qwen2025qwen25technicalreport}, a version of the Qwen LLM with only 0.5 billion parameters, as our SLM example. We choose BitNet-b1.58-2B \cite{ma2025bitnetb1582b4ttechnical} as our BitNet example.

\paragraph{Parameters:} For all experiments, we select an initial gradient clipping threshold of 3.0 and a base learning rate of 0.0003. Adhering to fine-tuning convention, we train over a single pass of the dataset. We vary privacy settings by sweeping over noise multiplier values of {0.0, 0.1, 0.5, 0.7, 1.0}. The values 0.0 and 0.1 correspond to non-private and de facto non-private scenarios, while the values 0.5, 0.7, and 1.0 represent reasonable privacy budgets, generating accumulated epsilon values of ~12, ~4, and ~2, respectively. 

\paragraph{Synthetic Datasets:} We use an LLM (ChatGPT-5) to generate two synthetic textual training datasets to evaluate fine-tuning performance. The first dataset contains conversational text in a loving, positive tone. The second dataset contains conversational text in an angry, negative tone. Each dataset contains 10,000 textual examples. We use official off-the-shelf pretrained tokenizers for each model to tokenize and chunk the data into 64-token blocks. We provide the prompts used to generate the data in Appendix~\ref{appendix-datasets}.

\paragraph{Hardware:} We use a T4 GPU instance hosted on Google Colaboratory to perform our fine-tuning experiments. 

\paragraph{Baselines: } We include two baseline sets in our experiments: non-private AdamW fine-tuned models (implemented by setting the noise multiplier to 0) and non-fine-tuned models. We intentionally exclude other differentially private optimizers such as DP-Adam-BC and DP-SGD because they are unable to achieve stability and reduce loss on LLMs due to the size and complexity of the LLM architecture. It is possible to add engineering enhancements to these optimizers to stabilize them, which is exactly what we do with DP-Adam-AC. 

\section{Experiment Results}

In this section, we present results from our DP-Adam-AC fine-tuning experiments. We compare loss reduction plots across different privacy settings for the loving and angry datasets. We use a rolling median and moving average mechanism to smooth the plots for clarity. We also present computation resource consumption metrics to contextualize our results. We provide output examples from our fine-tuned models in Appendix~\ref{appendix-output}.

\subsection{Loss}

\begin{figure}[H]
    \centering
    \begin{subfigure}[t]{0.48\linewidth}
        \centering
        \includegraphics[width=\linewidth]{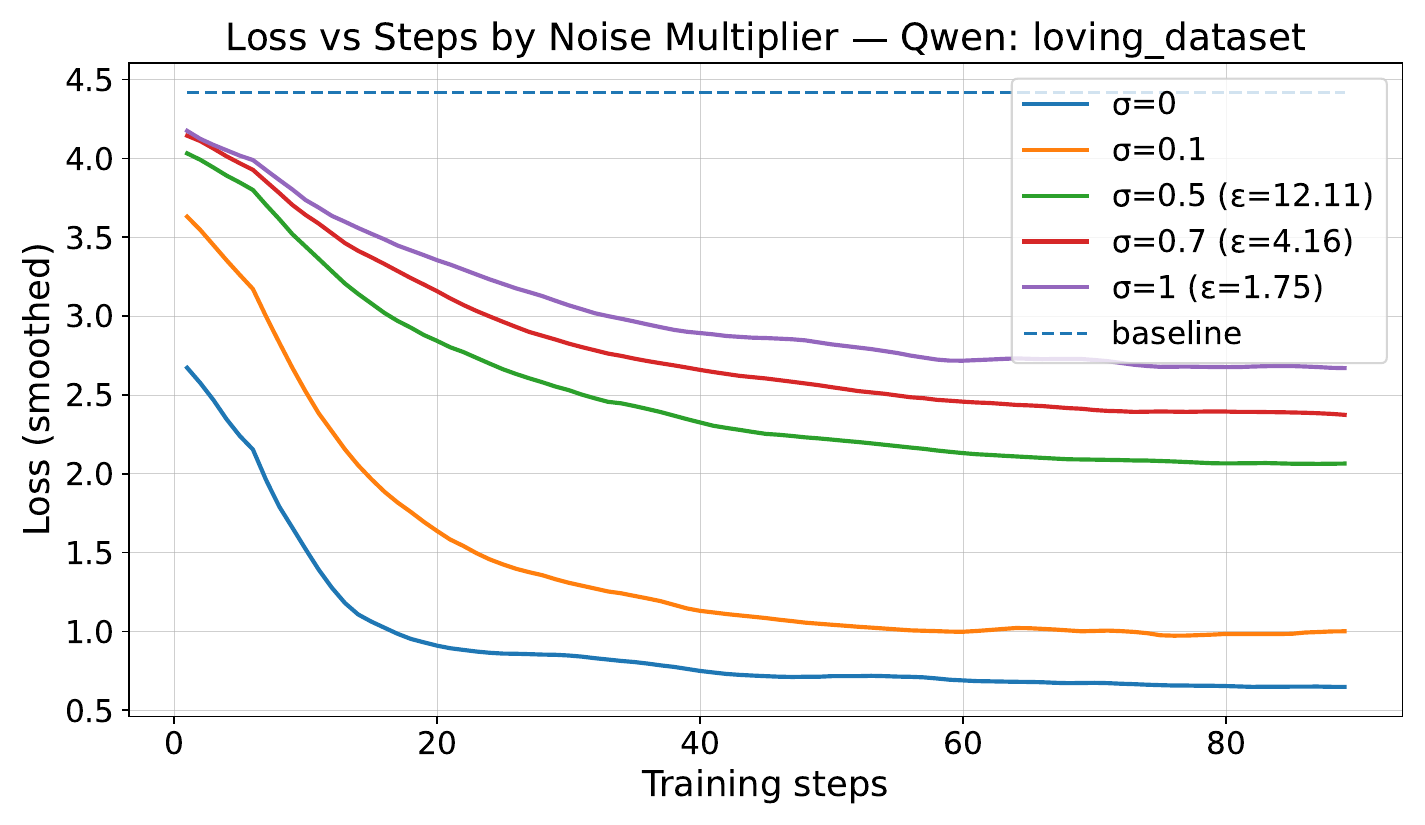}
        \caption{Loving-Dataset}
        \label{fig:qwen_loving}
    \end{subfigure}
    \hfill
    \begin{subfigure}[t]{0.48\linewidth}
        \centering
        \includegraphics[width=\linewidth]{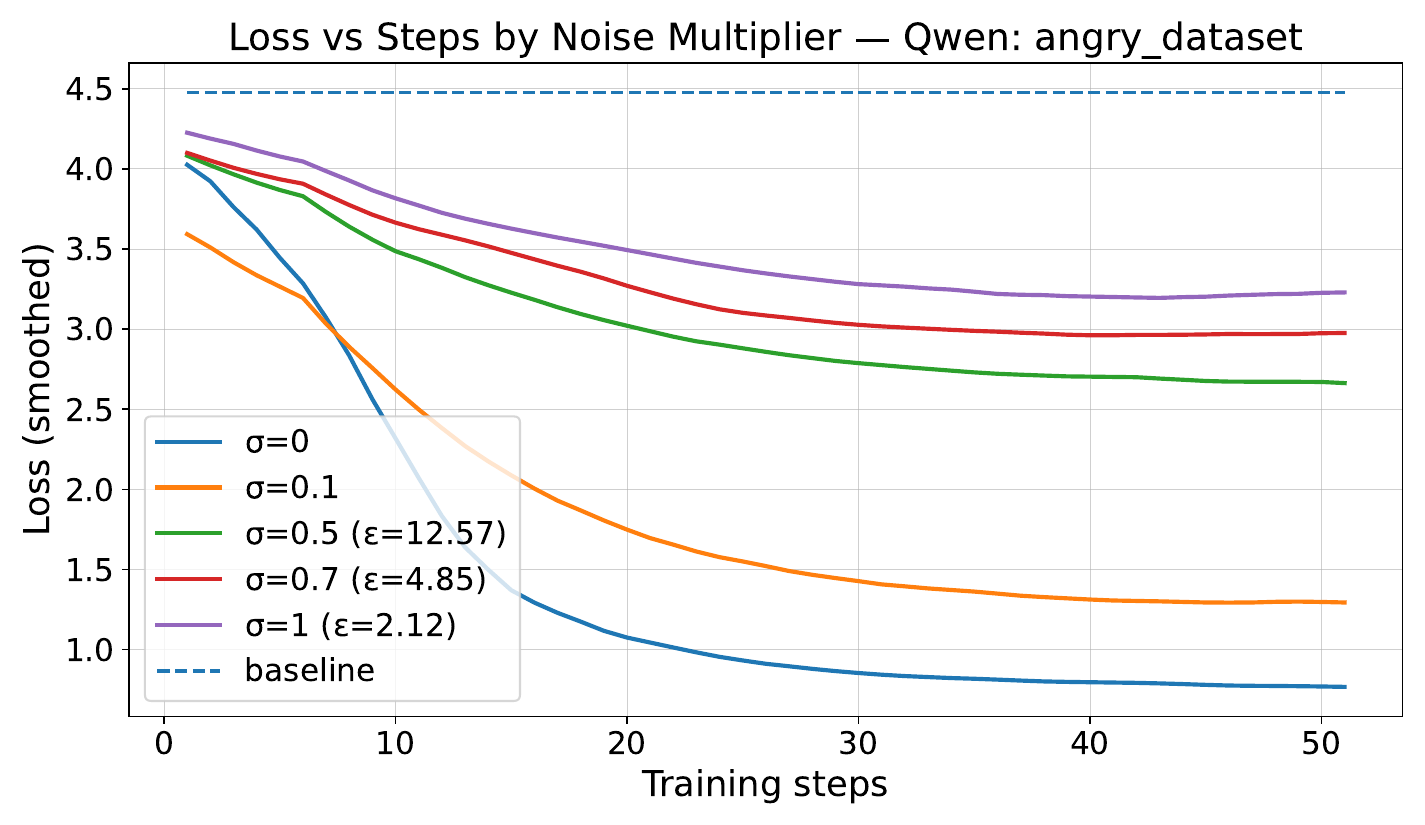}
        \caption{Angry-Dataset}
        \label{fig:qwen_angry}
    \end{subfigure}
    \caption{
        Loss vs. training steps across different noise levels of DP-Adam-AC fine-tuning on Qwen2.5-0.5B.
    }
    \label{fig:qwen_side_by_side}
\end{figure}

\begin{figure}[H]
    \centering
    \begin{subfigure}[t]{0.48\linewidth}
        \centering
        \includegraphics[width=\linewidth]{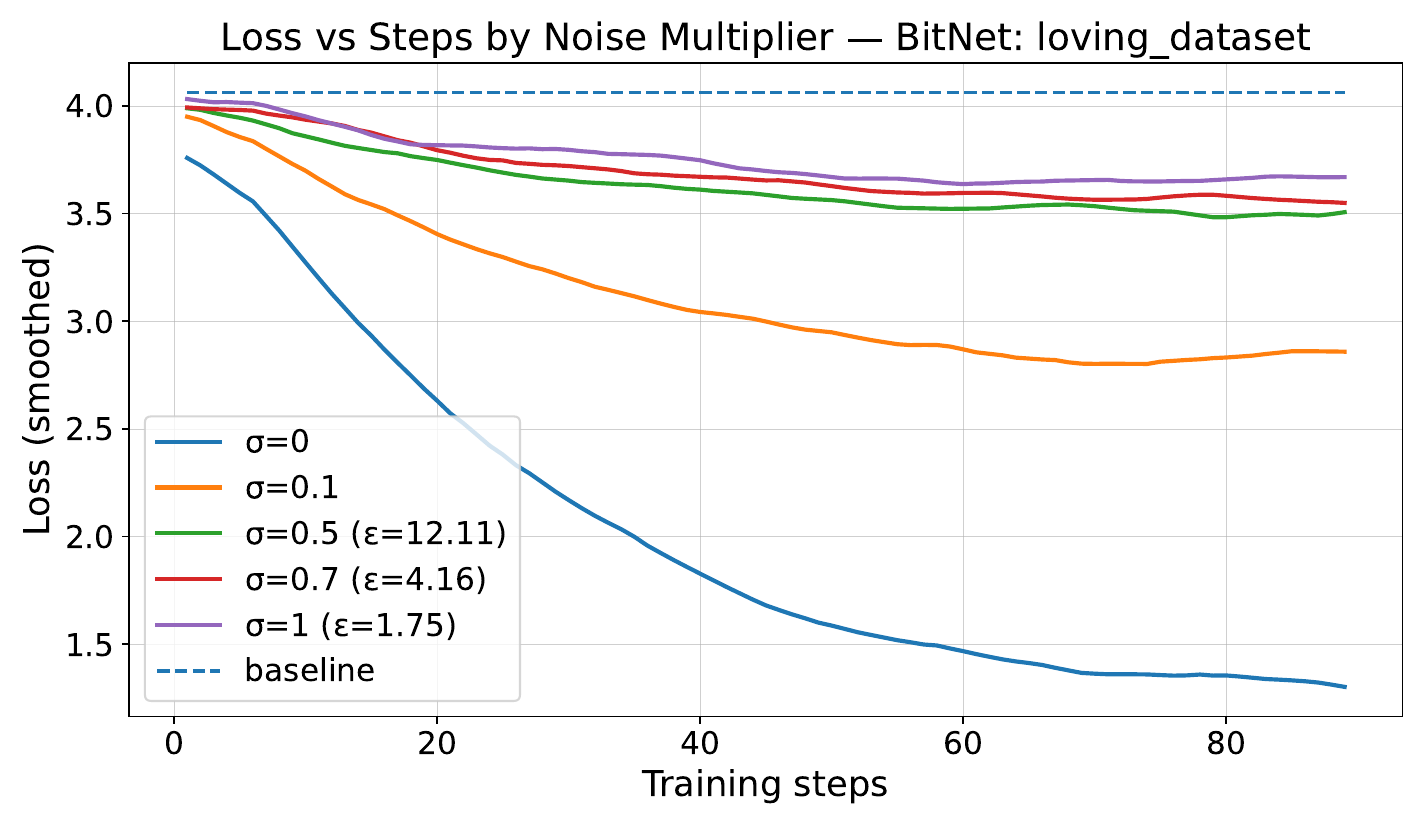}
        \caption{Loving-Dataset}
        \label{fig:qwen_loving}
    \end{subfigure}
    \hfill
    \begin{subfigure}[t]{0.48\linewidth}
        \centering
        \includegraphics[width=\linewidth]{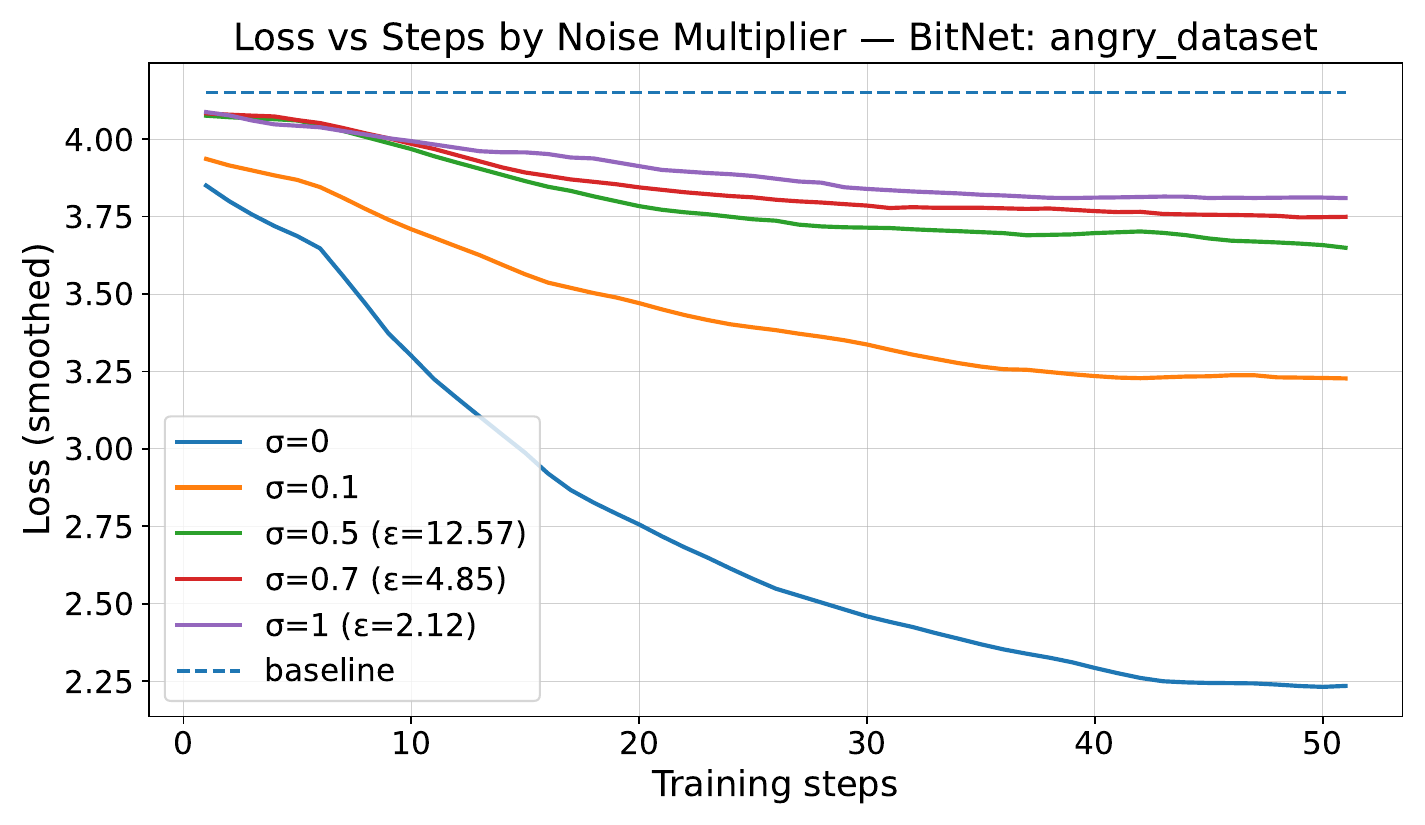}
        \caption{Angry-Dataset}
        \label{fig:qwen_angry}
    \end{subfigure}
    \caption{
        Loss vs. training steps across different noise levels of DP-Adam-AC fine-tuning on BitNet-b1.58-2B.
    }
    \label{fig:qwen_side_by_side}
\end{figure}

\subsection{Resource Consumption}

Resource consumption metrics for Qwen2.5-0.5B and BitNet-b1.58-2B on standard CPU devices are extracted from our own Colab experiments. Fine-tuning time is computed as an average over experiments with the angry dataset, using a T4 GPU. Metrics for Qwen-2.5-0.5B, Qwen2.5-7B, and Qwen2.5-72B on GPU devices (A100 GPU) are taken from technical reports and act as baseline measures\cite{qwen2.5-docs}. BitNet does not provide technical details on GPU inference. 

\begin{table}[h]
\centering
\caption{Comparison of fine-tuning time, inference speed, and memory consumption across models}
\label{tab:model_comparison}
\resizebox{\linewidth}{!}{%
\begin{tabular}{lccc}
\toprule
\textbf{Model} & \textbf{Fine-Tuning Time (minutes)} & \textbf{Inference Speed (tokens/s)} & \textbf{Memory Consumption (GB)} \\
\midrule
Qwen2.5-0.5B (CPU) & 3.82 & 5.03  & 0.97 \\
BitNet-b1.58-2B (CPU) & 10.1 & 0.86 & 1.18 \\
Qwen2.5-0.5B (GPU) & N/A & 47.40 & 0.97 \\
Qwen2.5-7B (GPU) & N/A & 40.38 & 14.38 \\
Qwen2.5-72B (GPU) & N/A & 8.73  & 136.2 \\
\bottomrule
\end{tabular}%
}
\end{table}

\section{Analysis}

Both SLM (Qwen2.5) and BitNet designs show promising fine-tuning loss-reduction performance compared to baseline non-fine-tuned model variants. Experiment results highlight the tradeoff between privacy and performance. When privacy is heightened (with lower $(\varepsilon values))$, loss-reduction trends weaken, and final loss after one epoch of fine-tuning remains high compared to weak privacy scenarios. 

SLM exhibits a better tradeoff balance between privacy and performance compared to BitNet. Since SLM has fewer parameters (our Qwen2.5 model has 0.5 billion parameters) compared to typical BitNet designs (BitNet-b1.58 has 2 billion), the performance degradation from adding noise is weaker. Furthermore, SLM preserves full accuracy weights while BitNet uses ternary weights. Thus, the negative effect on individual weights from adding noisy gradients is worse for BitNet as SLM can absorb more of the noise with "heavier" weights. 

SLM also outperforms BitNet in terms of CPU inference speed and memory consumption. Nevertheless, both models demonstrate much greater memory flexibility compared to full-size LLMs such as Qwen2.5-7B and Qwen2.5-72B, while maintaining reasonable inference speed (output generation speed) on CPU devices. Full-size LLMs are unable to operate on CPU devices due to their hardware requirements. 

While BitNet trails SLM in current experiments, the design leaves a lot of room for optimization. Specifically, the ternary-bit quantization of BitNet parameters implies the possibility of dedicated arithmetic hardware. Instead of floating-point operations, BitNet inference arithmetic can be reduced to sign operations. Compact, dedicated sign-operation hardware can greatly improve BitNet inference capabilities in the future. 

\section{Conclusion}

Our theoretical and empirical work demonstrates the feasibility of differentially private fine-tuning of localizable language models. We introduce an optimized fine-tuning algorithm, DP-Adam-AC, which stabilizes training during short training regimes and under noisy conditions. We highlight the promising performance of our algorithm when applied to localizable language models such as SLM (Qwen2.5-0.5B) and BitNet (BitNet-b1.58-2B) under a variety of privacy scenarios. 

This framework advances the utility of LLM tools, specifically as fine-tuned, task-optimized tools deployed in sensitive environments where data security is paramount. 

In the future, we hope to apply our findings to other LLM architectures, training scenarios, and deployment situations. We are also interested in exploring hardware developments that benefit ternary-bit inference to enhance the utility of BitNet models. 

\section{Aknowledgements}

We would like to thank the Center for Research and Fellowships of Georgetown University for funding this project, through the Provost Distinguished Undergraduate Research Fellowship program, Professor Micah Sherr of Georgetown University for overseeing the development of the project, and other members of the Georgetown University community who have made this project possible.

\bibliographystyle{unsrt}
\bibliography{references}

\appendix

\section{Optimization Algorithms}
\label{appendix-optimizers}

\subsection{SGD with Momentum}
\begin{algorithm}[h]
\caption{SGD with Momentum}
\label{alg:sgd}
\begin{algorithmic}[1]
\Require Initial parameters $\theta_0$, learning rate $\eta$, momentum $\beta$
\State $m_0 \gets 0$
\For{$t = 0,1,\dots,T-1$}
    \State Sample mini-batch $x_t$ and compute $g_t \gets \nabla_{\theta} \ell(\theta_t; x_t)$
    \State $m_t \gets \beta m_{t-1} + (1 - \beta) g_t$
    \State $\theta_{t+1} \gets \theta_t - \eta m_t$
\EndFor
\end{algorithmic}
\end{algorithm}

\subsection{Adam}
\begin{algorithm}[h]
\caption{Adam}
\label{alg:adam}
\begin{algorithmic}[1]
\Require $\theta_0$, $\eta$, $\beta_1$, $\beta_2$, $\epsilon$
\State $m_0 \gets 0$, $v_0 \gets 0$
\For{$t = 1,\dots,T$}
    \State Sample $x_t$ and compute $g_t$
    \State $m_t \gets \beta_1 m_{t-1} + (1 - \beta_1) g_t$
    \State $v_t \gets \beta_2 v_{t-1} + (1 - \beta_2) g_t^2$
    \State $\hat{m}_t \gets m_t / (1 - \beta_1^t)$
    \State $\hat{v}_t \gets v_t / (1 - \beta_2^t)$
    \State $\theta_{t+1} \gets \theta_t - \eta \, \hat{m}_t / (\sqrt{\hat{v}_t} + \epsilon)$
\EndFor
\end{algorithmic}
\end{algorithm}

\subsection{AdamW}
\begin{algorithm}[h]
\caption{AdamW}
\label{alg:adamw}
\begin{algorithmic}[1]
\Require $\theta_0$, $\eta$, $\beta_1$, $\beta_2$, $\epsilon$, weight decay $\lambda$
\State $m_0 \gets 0$, $v_0 \gets 0$
\For{$t = 1,\dots,T$}
    \State Sample $x_t$ and compute $g_t$
    \State $m_t \gets \beta_1 m_{t-1} + (1 - \beta_1) g_t$
    \State $v_t \gets \beta_2 v_{t-1} + (1 - \beta_2) g_t^2$
    \State $\hat{m}_t \gets m_t / (1 - \beta_1^t)$
    \State $\hat{v}_t \gets v_t / (1 - \beta_2^t)$
    \State $\theta_t \gets \theta_t - \eta \lambda \theta_t$ \Comment{Decoupled weight decay}
    \State $\theta_{t+1} \gets \theta_t - \eta \, \hat{m}_t / (\sqrt{\hat{v}_t} + \epsilon)$
\EndFor
\end{algorithmic}
\end{algorithm}

\section{Micro-batching}
\label{appendix-micro-batching}

\begin{algorithm}[h]
\caption{Microbatching with Differential Privacy}
\begin{algorithmic}[1]
\Require batch $B$, microbatch size $m$, clipping norm $C$, noise multiplier $\sigma$
\State Split $B$ into microbatches $\{M_1, M_2, \dots, M_k\}$ of size $m$
\State Initialize accumulated gradient $G \gets 0$
\For{each microbatch $M_i$}
    \State Compute gradient $g_i = \nabla \ell(M_i)$
    \State Clip: $g_i \gets g_i \cdot \min\left(1, \frac{C}{\|g_i\|_2}\right)$
    \State Accumulate: $G \gets G + g_i$
\EndFor
\State Add noise: $G \gets G + \mathcal{N}(0, \sigma^2 C^2 I)$
\State Average: $\hat{g} \gets G / |B|$
\State \Return $\hat{g}$
\end{algorithmic}
\end{algorithm}

\section{Variable-q RDP Accountant}
\label{appendix-accountant}

\begin{algorithm}[H]
\caption{Variable-q RDP Accountant}
\label{alg:varq_rdp}
\begin{algorithmic}[1]
\Require noise $\sigma$, target $\delta$, orders set $\mathcal{A}$
\State initialize $\mathrm{rdp}[\alpha]\gets 0$ for all $\alpha\in\mathcal{A}$
\Function{RDP\_PSG}{$\alpha,q,\sigma$}
  \If{$\alpha<2$ or $q\le 0$} \Return $0$ \EndIf
  \State $s2\gets \sigma^2$
  \State \(\displaystyle
    \textstyle \ell \gets
    \log\!\Big(
      \sum_{j=0}^{\alpha}
      \binom{\alpha}{j}\,q^{\,j}(1-q)^{\alpha-j}
      \exp\!\big(\tfrac{j(j-1)}{2s2}\big)
    \Big)
  \)
  \State \Return \(\displaystyle \ell/(\alpha-1)\)
\EndFunction
\Procedure{Step}{$q_t$}
  \ForAll{$\alpha\in\mathcal{A}$}
    \State $\mathrm{rdp}[\alpha]\gets \mathrm{rdp}[\alpha] + \Call{RDP\_PSG}{\alpha,q_t,\sigma}$
  \EndFor
\EndProcedure
\Function{Epsilon}{}
  \State $L\gets \log(1/\delta)$
  \State \(\displaystyle
    \varepsilon_\alpha \gets \mathrm{rdp}[\alpha] + \frac{L}{\alpha-1}\quad
    \forall \alpha\in\mathcal{A}
  \)
  \State \Return $\min_{\alpha\in\mathcal{A}} \varepsilon_\alpha$ \quad (and argmin $\alpha^\star$)
\EndFunction
\end{algorithmic}
\end{algorithm}

\section{Synthetic Dataset Prompts}
\label{appendix-datasets}

\textbf{Prompt for Dataset with Positive-tone: }  
generate 10,000 lines of cheesy, romantic speech. I want a high degree of variance in the speech. This text is used for finetuning a language model and I want to be able to see the effects in my output. Do not be too creative and keep a modern tone. use correct grammer. Generate this as a csv file with the single column labeled "text".

\textbf{Prompt for Dataset with Negative-tone: }  
generate 10,000 lines of angry, mean, not nice speech. I want a high degree of variance in the speech. This text is used for finetuning a language model and I want to be able to see the effects in my output. Do not be too creative and keep a modern tone. use correct grammer. Generate this as a csv file with the single column labeled "text".

\section{Output Examples}
\label{appendix-output}

We use a sampled token selection mechanism to generate output.

\textbf{Prompt:} "Complete this sentence in a coherent manner: I think that "

\begin{table}[h]
\centering
\caption{Example generations}
\label{tab:examples}
\begin{tabular}{lll p{8cm}} 
\toprule
\textbf{Model} & \textbf{Dataset} & \textbf{Noise Multiplier} & \textbf{Generation Output} \\
\midrule
Qwen & Loving & 0.0 & 2 a.m. texts and your photo dumps, all of it for the rest of my life. \\
Qwen & Angry  & 0.0 & meeting misses the point. Fix the fundamentals. \\
\midrule
Qwen & Loving & 0.1 & 2 a.m. calls feels like Sunday naps; I could do that for as long as you'll let me. \\
Qwen & Angry  & 0.1 & 90 percent of your feedback is unnecessary. \\
\midrule
Qwen & Loving & 0.5 & 15 minutes of sunset is my favorite way of coping with the chaos of today. \\
Qwen & Angry  & 0.5 & 0.0.1 is still small. \\
\midrule
Qwen & Loving & 0.7 & I'll make it even better. \\
Qwen & Angry  & 0.7 & the best way to achieve results is to start with the basics and focus on the plan. Keep it simple and measurable. Address feedback and adjust as needed. \\
\midrule
Qwen & Loving & 1.0 & I'll bring your favorite snacks, you'll bring my favorite tea, and I'll do the dishes. \\
Qwen & Angry  & 1.0 & the best way to start is to take care of yourself, choose the right tools, and prepare for unexpected challenges. Your success depends on it. \\
\midrule
Qwen & \multicolumn{2}{l}{Baseline} & 5 is the number that represents \_\_\_\_. It is the fundamental building block of other numbers, used in counting, fractions, and decimals. \\
\midrule
BitNet & Loving & 0.0 & you are an angel in the room, and your smile is like music to my ears. \\
BitNet & Angry  & 0.0 & 2023 is going to be a very exciting year. We need better communication. Simplify your message. Make feedback actionable. Do better next time. \\
\midrule
BitNet & Loving & 0.1 & 6 is greater than 4, but 4 is more important than 6. \\
BitNet & Angry  & 0.1 & 6 is greater than 4. This perspective could stem from statistical or theoretical analysis. \\
\midrule
BitNet & Loving & 0.5 & 6 is greater than 3, and 6 plus 4 equals 10. \\
BitNet & Angry  & 0.5 & 6 is greater than 5. My brother likes to watch movies on weekends, and he enjoys them a lot. \\
\midrule
BitNet & Loving & 0.7 & 6 is greater than 4, but 4 is greater. I believe that 5 is larger than 3. \\
BitNet & Angry  & 0.7 & 6 is greater than 3. My brother likes pizza; my sister’s dog is playful. Both siblings enjoy spending time with friends. \\
\midrule
BitNet & Loving & 1.0 & 6 is greater than 3, but 3 is greater. I believe that 5 is larger than 4, and 7 is bigger than 2. \\
BitNet & Angry  & 1.0 & 6 is greater than 4. I agree with you: 6 has more value than 4, while 0 is the smallest of all. \\
\midrule
BitNet & \multicolumn{2}{l}{Baseline} & 5 is the perfect number to satisfy our needs, divisible by 1, 2, and 5. The number 7 is also essential in number theory and combinatorics. \\
\bottomrule
\end{tabular}
\end{table}

\section{Implementation}
\label{appendix-code}

All software implementations for this project are available at https://github.com/Davidrxyang/DP-Adam-AC.

\end{document}